\PassOptionsToPackage{type={CC}, modifier={by}, version={4.0}}{doclicense}
\documentclass[sigconf, nonacm]{acmart}
\setcopyright{none}
\newcommand{\ourplatform}{Deep Research Comparator\xspace}
\newcommand{\oursys}{Simple Deepresearch\xspace}
\usepackage{makecell}

\usepackage{listings}

\lstdefinelanguage{json}{
    string=[s]{"}{"},
    comment=[l]{//},
    morecomment=[s]{/*}{*/},
    literate=
        *{0}{{{\color{blue}0}}}{1}
         {1}{{{\color{blue}1}}}{1}
         {2}{{{\color{blue}2}}}{1}
         {3}{{{\color{blue}3}}}{1}
         {4}{{{\color{blue}4}}}{1}
         {5}{{{\color{blue}5}}}{1}
         {6}{{{\color{blue}6}}}{1}
         {7}{{{\color{blue}7}}}{1}
         {8}{{{\color{blue}8}}}{1}
         {9}{{{\color{blue}9}}}{1}
}
\definecolor{softgreen}{RGB}{34,139,34} 

\newcommand{\cmark}{\textcolor{green!60!black}{\ding{51}}}
\newcommand{\xmark}{\textcolor{red}{\ding{55}}}

\usepackage{enumitem}

\usepackage[utf8]{inputenc}

\usepackage{pifont}     
\usepackage{xspace}

\usepackage{multirow}
\usepackage{stfloats}

\AtBeginDocument{%
  }

\setcopyright{acmlicensed}
\copyrightyear{2026}
\acmYear{2026}
\acmDOI{}
\acmConference[WWW '26]{The Web Conference 2026}{April 13--17, 2026}{Dubai, United Arab Emirates}
\acmBooktitle{Companion Proceedings of the ACM Web Conference 2026 (WWW '26 Companion)}

\begin{document}

\title{Deep Research Comparator: A Platform For Fine-grained Human Annotations of Deep Research Agents}

\author{Prahaladh Chandrahasan}
\authornote{These authors contributed equally to this work.}
\email{prahalac@andrew.cmu.edu}
\affiliation{%
  \institution{Carnegie Mellon University}
  \city{Pittsburgh}
  \state{Pennsylvania}
  \country{USA}
}

\author{Jiahe Jin}
\authornotemark[1]
\email{jjiahe@andrew.cmu.edu}
\affiliation{%
  \institution{Carnegie Mellon University}
  \city{Pittsburgh}
  \state{Pennsylvania}
  \country{USA}
}

\author{Zhihan Zhang}
\authornotemark[1]
\email{zhihanz@andrew.cmu.edu}
\affiliation{%
  \institution{Carnegie Mellon University}
  \city{Pittsburgh}
  \state{Pennsylvania}
  \country{USA}
}

\author{Tevin Wang}
\email{tevinw@andrew.cmu.edu}
\affiliation{%
  \institution{Carnegie Mellon University}
  \city{Pittsburgh}
  \state{Pennsylvania}
  \country{USA}
}

\author{Andy Tang}
\email{andyt@andrew.cmu.edu}
\affiliation{%
  \institution{Carnegie Mellon University}
  \city{Pittsburgh}
  \state{Pennsylvania}
  \country{USA}
}

\author{Lucy Mo}
\email{linmo@andrew.cmu.edu}
\affiliation{%
  \institution{Carnegie Mellon University}
  \city{Pittsburgh}
  \state{Pennsylvania}
  \country{USA}
}

\author{Morteza Ziyadi}
\email{mziyadi@amazon.com}
\affiliation{%
  \institution{Amazon}
  \city{Seattle}
  \state{Washington}
  \country{USA}
}

\author{Leonardo F.R. Ribeiro}
\email{leonribe@amazon.com}
\affiliation{%
  \institution{Amazon}
  \city{Seattle}
  \state{Washington}
  \country{USA}
}

\author{Zimeng Qiu}
\email{zimengqi@amazon.com}
\affiliation{%
  \institution{Amazon}
  \city{Seattle}
  \state{Washington}
  \country{USA}
}

\author{Markus Dreyer}
\email{mddreyer@amazon.com}
\affiliation{%
  \institution{Amazon}
  \city{Seattle}
  \state{Washington}
  \country{USA}
}

\author{Akari Asai}
\email{aasai@cs.cmu.edu}
\affiliation{%
  \institution{Carnegie Mellon University}
  \city{Pittsburgh}
  \state{Pennsylvania}
  \country{USA}
}

\author{Chenyan Xiong}
\email{cx@cs.cmu.edu}
\affiliation{%
  \institution{Carnegie Mellon University}
  \city{Pittsburgh}
  \state{Pennsylvania}
  \country{USA}
}

\renewcommand{\shortauthors}{Chandrahasan et al.}


\renewcommand{\thefootnote}{\arabic{footnote}}
\begin{abstract}
Evaluating deep research agents that iteratively search the web, analyze information, and generate reports remains a major challenge, especially in assessing long reports and providing fine-grained feedback.
To address these gaps, we introduce \textsc{\ourplatform}, a holistic annotation platform for the human evaluation of deep research agents. Our platform displays the generated reports and intermediate steps from two agents side-by-side, and allows annotators to state their preference between two final reports, and provide fine-grained feedback on specific text spans within the report or intermediate steps for each agent separately.
Furthermore, we develop \textsc{\oursys}, an agent scaffold that serves as a baseline to facilitate the integration of various large language models to transform them into deep research agents for evaluation. Experiment results of three agents with 17 annotators demonstrated strong alignment between human annotations and static benchmarks, while revealing divergence in fine-grained metrics, which highlights the distinct value of fine-grained annotations in providing robust evaluation and actionable data for nuanced agent optimization. We will provide a web-based implementation of our platform that conference attendees can interact with in real-time. \footnote{Code and video demo are available at \url{https://github.com/cxcscmu/Deep-Research-Comparator}}.

\end{abstract}



\received{17 November 2025}


\maketitle

\section{Introduction}
Recent advancements in large language models (LLMs) have catalyzed the development of agentic search, a new paradigm in information retrieval. Within this paradigm, \textit{deep research agents} have emerged as LLM-powered systems that autonomously find, analyze, and synthesize vast online information to produce comprehensive reports. Prominent commercial systems from OpenAI and Google have gained wide adoption, and the open-source community too has produced notable contributions~\cite{li2025webthinkerempoweringlargereasoning, GPTResearcher}.

\begin{figure*}[h]
  \centering
  \includegraphics[width=0.95\textwidth]{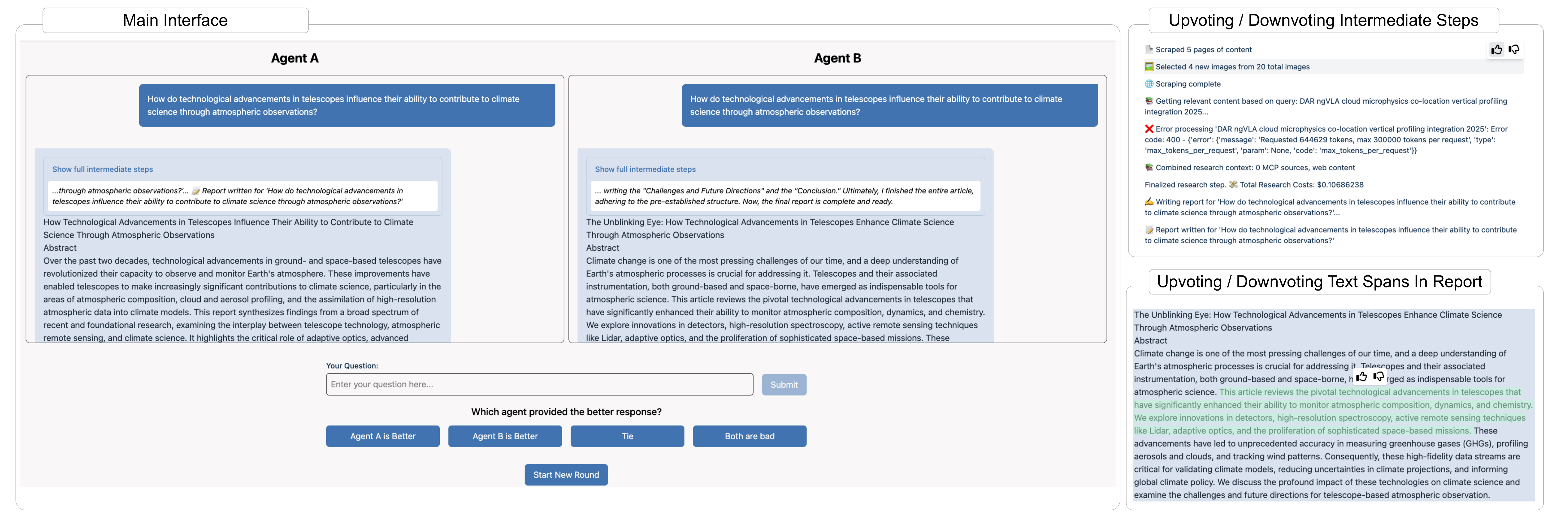}
  \caption{Left: the main interface of \textsc{\ourplatform}. Right: the fine-grained annotation mechanism for upvoting/ downvoting intermediate steps and text spans in report.}
  \label{fig:interface}
\end{figure*}

Despite these advancements, evaluating deep research agents remains a significant challenge. Existing benchmarks primarily target verifiable short-form answers, leaving the assessment of long-form reports underexplored. While some works use hand-crafted rubrics for evaluation~\cite{gou2025mind2web2evaluatingagentic}, this approach is static and hard to generalize to new tasks or domains. 

A key challenge in evaluating long report generation of deep research agents is the lack of fine-grained assessment. Current methods often assess reports holistically~\cite{li2025webthinkerempoweringlargereasoning, du2025deepresearch}, failing to capture the uneven quality within the report, such as strong insights mixed with inaccurate content. This lack of text span level feedback makes it difficult to pinpoint specific areas for improvement and perform targeted agent optimization. Furthermore, there is a lack of process-level evaluation for the multi-step actions that agents execute. Current evaluations are typically based on outcomes, resulting in sparse and delayed feedback. However, process feedback is crucial for training methods like reinforcement learning (RL) for accurate contribution attribution. Such feedback is also essential for understanding agent behavior and assessing the reliability of the generated reports.

To address these gaps, we introduce \textsc{\ourplatform}, a platform designed to facilitate holistic human annotation for deep research agents. It supports easy integration of various agents with minimal engineering effort, enabling any agent system to be hosted through a unified web interface. The platform randomly routes user queries to two agents and presents their intermediate steps and final reports. 
\textsc{\ourplatform} enables the collection of two types of human feedback (Figure~\ref{fig:interface}). First, for overall report evaluation, it collects preference votes via side-by-side comparisons of final reports, similar to Chatbot Arena~\cite{chiang2024chatbot}. Second, for fine-grained evaluation, it supports upvoting or downvoting individual intermediate steps in the generation process and specific text spans in the final reports.

To support the evaluation of different LLMs as deep research agents, 
we developed \textsc{\oursys}, an agent scaffold serving as a baseline, which allows various LLMs to be easily integrated to transform them into deep research agents for evaluation.

Using our platform, we conducted a proof-of-concept experiment on three agents with 176 user queries, collecting preference votes on overall report quality and upvote/downvote feedback on 1,281 intermediate steps and 593 text spans. Results show human preference votes on overall report quality closely match static-benchmark evaluations, confirming the platform’s reliability. At the same time, the divergence in fine-grained metrics demonstrates these signals capture aspects of agent behavior that holistic scores miss. By providing a platform for collecting such rich fine-grained annotations, we aim to enable more robust evaluation, deeper diagnostics, and targeted optimization of deep research agents.

Overall, we make three key contributions:
\begin{enumerate}
\item We provide \textsc{\ourplatform}, an open source platform to host deep research agents and supports fine-grained human annotation.
\item We develop \textsc{\oursys}, a baseline agent scaffold that allows for the easy integration of various LLMs to transform them into deep research agents.
\item We validate \textsc{\ourplatform} via a annotation experiment showing close alignment with static benchmarks on overall report quality and meaningful divergence in fine-grained metrics, collected from 176 real-world queries with 1,281 step and 593 span annotations.
\end{enumerate}

\section{Related Work}
Search agents are LLM-based systems that autonomously use search tools to gather information~\cite{jin2025searchr1trainingllmsreason, schick2023toolformerlanguagemodelsteach, claudesearch, chen2024mindsearch} for solving complex search tasks. These agents can execute real-time internet searches, and synthesize information to provide responses. 
Among them, ``deep research agents'' represent a more capable evolution~\cite{OpenAIdeepresearch, GPTResearcher, li2025webthinkerempoweringlargereasoning}, which conduct substantially more searches, engage in long-horizon planning, and operate with significantly higher test-time compute budgets, for more research-intensive scenarios. 

Evaluations for search agents fall into two types: short-answer tasks and long-form report generation. Short-answer tasks are easier to evaluate due to definitive ground truth. These include tasks demanding extensive knowledge and complex reasoning ~\cite{rein2023gpqagraduatelevelgoogleproofqa, phan2025humanitysexam}, and specialized web search benchmarks~\cite{mialon2023gaiabenchmarkgeneralai, wei2025browsecompsimplechallengingbenchmark, futuresearch2025deepresearchbenchevaluating,wu2025webwalkerbenchmarkingllmsweb, wei2025browsecompsimplechallengingbenchmark}. In contrast, evaluating long-form reports is more challenging. Some efforts introduced hand-crafted, rubric-based benchmarks~\cite{gou2025mind2web2evaluatingagentic, arora2025healthbenchevaluatinglargelanguage,futuresearch2025deepresearchbenchevaluating,asai2024openscholarsynthesizingscientificliterature}, while other studies explored using LLMs to automatically score reports against predefined criteria~\cite{coelho2025deepresearchgymfreetransparentreproducible, li2025webthinkerempoweringlargereasoning}. As summarized in Table~\ref{tab:dataset_comparison}, \ourplatform is distinguished from prior work by supporting evaluation on real-world user queries, long-form reports, and fine-grained feedback mechanisms.

\section{\textsc{\ourplatform}}
\label{sec: deep_research_com}

This section presents \textsc{\ourplatform}, a platform for fine-grained human annotation of deep research agents.

\subsection{User Interface and Evaluation Workflow}
\label{sec:ui}

Our platform provides a web interface for side-by-side comparison and fine-grained annotation, as shown in Figure~\ref{fig:interface}. Annotators can submit arbitrary queries, and our system will assign the query to two anonymous deep research agents. As the report generation begins, the agents' intermediate steps are streamed to the front-end, followed by the final reports. After both reports are generated, users vote for their preferred report by selecting one of four choices: Agent A, Agent B, Tie, or Both are bad. These pairwise comparisons are used to compute a ranking for overall report preference.

Besides evaluating the overall quality of final reports, this platform also supports fine-grained annotation. Each step in the agents' generation process is accompanied by upvote and downvote icons, allowing users to provide step-wise feedback. As a long report may contain content of varying quality, our platform also supports highlighting specific text spans in the final report and providing feedback as upvotes or downvotes.

\subsection{System Architecture}
\label{sec:system_architecture}

\begin{figure*}[h]
  \centering
  \includegraphics[width=0.8\textwidth]{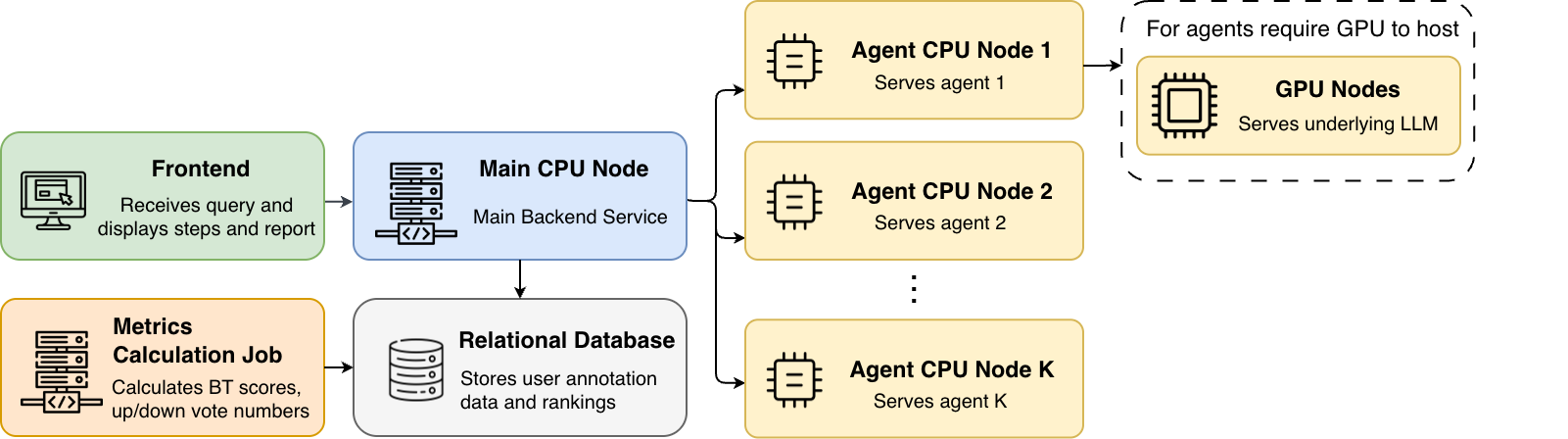}
  \caption{High-level overview of \textsc{\ourplatform}'s system architecture.}
  \label{fig:system-architecture}
\end{figure*}

Figure~\ref{fig:system-architecture} presents an overview of our architecture. Our platform supports hosting a wide range of deep research agents. Agents that generate intermediate steps and final report via a unified JSON interface can be integrated with minimal engineering effort. This design allows developers to flexibly customize the content of intermediate steps for their annotation requirements and imposes no constraints on the agent’s internal architecture.

\paragraph{Frontend}
The frontend presents generated intermediate steps and final reports on the user interface. Implemented as a static web application, its static files are produced during the build process and hosted in an AWS S3 bucket, served through CloudFront CDN caching ensuring low latency.

\paragraph{Main Backend Service}
The system routes user queries from the frontend to the Main Backend Service hosted on the main CPU node. It handles user requests, selects agents, routes queries, processes user votes, and serves intermediate steps, final reports, and rankings back to the frontend.

\paragraph{Agent Serving Service}
User queries are routed to the Agent Serving Services deployed on dedicated agent CPU nodes, each encapsulated within a Docker container. These agent services stream their responses back to the main backend via internal API calls using the unified JSON format. For open-source agent services whose underlying LLMs require GPU resources, the LLMs are deployed on specialized GPU nodes and accessed by the agent services through internal API calls. For closed-source agents, specialized proxy pods are implemented to invoke the agents through external APIs.

\paragraph{Metrics Calculation Job}
This CronJob reads agent conversations and user annotation data from the database, computing outcome-based evaluation metrics (like report-preference rankings) and fine-grained metrics (such as upvote rate). It then writes the results back to the relational database, which are served to the user interface via the Main Backend Service.

\section{\textsc{\oursys}}
\label{sec: baseline scaffold}

\textsc{\oursys} is a prompt-based, end-to-end agent scaffold that allows for the easy integration of different LLMs via the OpenAI API, transforming it into a deep research agent. It serves as a baseline for evaluation.

The agent's workflow is an iterative process. At step $k$, the agent uses the user query $q$ and the history context $ctx_k$ as input, and generates the thought process $t_k$ and action $a_k$. Each action consists of action type and corresponding content, adhering to a predefined action space (see Table~\ref{tab:action_space}) specified in the agent's system prompt. If $a_k$ is a \texttt{search} action, the agent queries a search API and receives documents as observation $obs_k$. Other actions update the agent's memory by being incorporated into the history of next step.

The \texttt{summary} action is introduced to better accommodate models with shorter context lengths. This allows the model to summarize the entire history, preserving the information it considers most important. The content of \texttt{summary} action will be used as history for the next step.

\begin{table}[!ht]
\centering
\caption{The action space of \textsc{\oursys}.\vspace{-3ex}}
\label{tab:action_space}
\begin{tabular}{ll}
\toprule
\textbf{Type} & \textbf{Content} \\
\midrule
\texttt{plan} & The plan for the research process. \\
\texttt{search} & Web search query. \\
\texttt{script} & Draft for the final report. \\
\texttt{summary} & Summary of history steps. \\
\texttt{answer} & The final report. \\
\bottomrule
\end{tabular}
\end{table}

\vspace{-1.3em}

\section{Experiments}
\label{sec: experiments}
\begin{table*}[h!]
  \centering
  \caption{Evaluation results of deep research agents: human evaluation on \textsc{\ourplatform}(Ranking, BT Score, Step Upvote Rate, Text Span Upvote Rate) and LLM-based evaluation on DeepResearchGym (Clarity, Insight).}
  \label{tab:main_results}
  \renewcommand{\arraystretch}{1}
  \resizebox{\textwidth}{!}{
  \begin{tabular}{l|cc|cccc}
    \toprule
    & \multicolumn{2}{c|}{\textbf{Static Benchmark Evaluation}} 
    & \multicolumn{4}{c}{\textbf{Human Annotation on \ourplatform}} \\
    \cmidrule(lr){2-3} \cmidrule(lr){4-7}
    \textbf{Agent} & 
    \textbf{Clarity} & 
    \textbf{Insight} &
    \textbf{Ranking} & 
    \textbf{BT Score} & 
    \makecell{\textbf{Step Upvote} \\ \textbf{Rate (\%)}} & 
    \makecell{\textbf{Text Span Upvote} \\ \textbf{Rate (\%)}} \\
    \midrule
    GPT Researcher              
    & \textbf{89.8} & \textbf{90.4} 
    & \textbf{1} & \textbf{1135.28} & 88.59 & 74.66 \\

    Perplexity DeepResearch    
    & 89.5 & 89.3
    & 2 & 1087.41 & \textbf{90.44} & \textbf{90.72} \\

    \textsc{\oursys} (Gemini 2.5 Flash) 
    & 89.4 & 79.4
    & 3 & 1000.00 & 78.91 & 90.33 \\
    
    \bottomrule
  \end{tabular}
  }
\end{table*}

\subsection{Human Annotation}

We conduct human annotation experiments as proof-of-concept, with setup and evaluation details provided below.

\paragraph{Agents}
We evaluated three deep research agents: (1) the proprietary system \textbf{Perplexity Deep Research}, (2) the open-source ~\textbf{GPT Researcher}~\cite{GPTResearcher}, configured in deep research mode with GPT-4.1 as the smart LLM, o4-mini as the strategic LLM, and Serper as the search API (3) and \textbf{\textsc{\oursys} (Gemini 2.5 Flash)}, our baseline agent scaffold integrated with Gemini 2.5 Flash as the underlying LLM and uses ClueWeb22 as search API.

\paragraph{Annotators}
A total of 17 university students and industry researchers participated in the data annotation process. Annotators were instructed to submit their own queries to the platform that reflect significant information needs.

\paragraph{Metrics}

We present two primary types of metrics computed from the platform's annotations. For the evaluation of overall report quality, we follow Chatbot Arena~\cite{chiang2024chatbot} to use the Bradley-Terry (BT) model~\cite{bradley1952rank} to compute BT scores from pairwise preference votes. These scores are computed relative to our baseline \textsc{\oursys}, which is assigned a fixed score of 1000. We then derive a report-preference ranking from these scores as assessment for overall report quality. We also introduce a fine-grained evaluation metric: the \textit{upvote rate}. For each agent, we aggregate all upvotes and downvotes received by its intermediate steps and text spans and compute the \textit{upvote rate} as $\text{upvotes} / (\text{upvotes} + \text{downvotes})$. It is important to note that these metrics are representative examples; \textsc{\ourplatform} collects rich raw data (including pairwise preferences, step-wise votes, and text-span votes) that researchers can export and utilize for their own custom analyses. 

\subsection{Evaluation on Static Benchmark}

To compare human evaluation for overall report quality with those from static benchmarks, we evaluated agents on DeepResearchGym~\cite{coelho2025deepresearchgymfreetransparentreproducible}, a benchmark that uses LLM-as-a-judge to assess reports generated by deep research agents. 

Our evaluation employs two metrics targeting different aspects of the overall report quality. We provide the LLM judge with detailed evaluation criteria and prompt it to assign scores based on it. Specifically, \textit{Clarity} assesses the logical coherence and linguistic fluency of the report, while \textit{Insight} captures the analytical nuance and depth of reasoning demonstrated.

\subsection{Results}
The annotation experiment yielded 176 queries across a diverse range of domains---including science, technology, society, economics, health, and culture. The resulting BT scores and the report-preference ranking are presented in Table~\ref{tab:main_results}. In addition to preference votes, we collected 1,281 upvote/downvote feedback instances on intermediate steps and 593 on highlighted text spans within the reports. The corresponding upvote rates are shown in Table~\ref{tab:main_results}.

We observe a strong alignment between our human evaluation results with static benchmark evaluation. As shown in the first four columns, the scores from the static benchmark perfectly matches the human-derived report-preference ranking. However, there's a divergence between this overall quality based ranking and the fine-grained metric upvote rates, as shown in the last four columns.


\subsection{Discussion}
The strong alignment between our human annotation and static benchmark on overall report quality validates the platform's reliability. More importantly, the divergence between this overall quality and the fine-grained metrics highlights that \textsc{\ourplatform} provides a more nuanced perspective for agent development. The fine-grained feedback offers actionable data beyond a single outcome score. For example, text-span feedback can guide fine-grained optimization of report content, while step-wise feedback provides a valuable resource for training a process reward model or data synthesis for offline reinforcement learning.

\section{Conclusion}
We present an open-source platform that offers a framework for holistic annotation for deep research agents. We also develop a baseline agent scaffold that allows the integration of different LLMs to evaluate them as deep research agents. Using our platform, we collected votes and fine-grained annotation for 176 user queries. We believe \textsc{\ourplatform} will serve as a useful resource for community by benefiting a wide range of downstream applications, such as benchmarking, agent behavior analysis, and agent training via techniques like process supervision and RLHF. 

\section{Ethics Statement}
All annotators were members of our research group who provided 
informed consent before contributing anonymous annotations, with the 
right to withdraw at any time. No personally identifiable information 
was collected or retained during the study.
\bibliographystyle{ACM-Reference-Format}
\bibliography{main}

\appendix
\clearpage

\section{\textsc{\oursys} Scaffold Prompt}
\label{sec:prompt_appendix}

\begin{lstlisting}
You are a research assistant with the ability to perform web searches to write a comprehensive scientific research article in markdown format. You will be given a question, and you will need to write a report on the question. You can use search tools to find relevant information.
You don't need to write the report in one turn. You can search and revise your report multiple times. When you consider you need some new information, you can perform a search action. When you want to update, generate, or revise your report scripts, you can perform a scripts action. When you consider you have enough information, you can output the final report.

Based on the history information, you need to suggest the next action to complete the task. 
You will be provided with:
1. Your history turns information: it might contains your previous plan, report scripts, search results. For search results, queries are in format <search> query </search> and the returned search results in <information> and </information>.
2. The question to answer.

IMPORTANT: You must strictly adhere to the following rules:
1. Choose ONLY ONE action from the list below for each response, DO NOT perform more than one action per step.
2. Follow the exact syntax format for the selected action, DO NOT create or use any actions other than those listed.
3. **Don't do duplicate search.** Pay attention to the history search results.
4. **Do not always perform the search action. You must consider the history search results and update your report scripts.**

Valid actions:
1. <search> query </search>: search the web for information if you consider you lack some knowledge.
2. <plan> plan </plan>: plan the report in your first turn.
3. <scripts> revised or newly generated report scripts </scripts>: revise former report scripts, or newly generate report scripts.
4. <summary> important parts of the history turns </summary>: summarize the history turns. Reflect the plan, scripts, search queries, and search results in you history turns, and keep the information you consider important for answering the question and generating your report. Still keep the tag structure, keep plan between <plan> and </plan>, keep scripts between <scripts> and </scripts>, keep search queries between <search> and </search>, and keep search results between <information> and </information>. The history turn information for your subsequent turns will be updated accoring to this summary action.
5. <answer> final report </answer>: output the final report.

Question: {question}

History Turns: (empty if this is the first turn)

\end{lstlisting}

\section{History Context Update Mechanism of \textsc{\oursys}}
\label{app: history}

The history context at step $k$, $ctx_k$, is part of the agent's input that contains the information of the preceding steps (from 1 to $k-1$), providing the agent with the necessary information for its current step. If $a_k$ is an \texttt{answer} action, then the generation process ends. Otherwise, the history context for step $k+1$ is updated as follows:

{\small
\[
ctx_{k+1} =
\begin{cases}
  ctx_k + t_k + a_k + obs_k, & a_k = \texttt{Search} \\
  ctx_k + t_k + a_k, & a_k = \texttt{Plan, Script} \\
  ctx_k, & a_k = \texttt{Summary}
\end{cases}
\]
}

\section{Unified JSON format}
\label{sec:JSON_schema}
\begin{lstlisting}[language=JSON ]
# Standardized fields that all agents must support:
{
    "intermediate_steps": str, 
    "final_report": str, 
    "is_intermediate": bool, 
    "is_complete": bool, 
    "citations": List[str] 
} + "|||---|||"

\end{lstlisting}

\section{Annotation Example}
\label{sec: annotation}

This is a annotation example from our human annotation experiment:
\newline
\noindent\textbf{Question:} \textit{How does the inclusion of personal life details in journalist biographies affect public perception of their professional credibility and integrity?}

\noindent\textbf{Session ID:} \texttt{24bed070-c202-46f3-ba23-37cf42b88f59} \\
\noindent\textbf{Agent A:} \textsc{\oursys} (Gemini 2.5 Flash) \\
\noindent\textbf{Agent B:} Perplexity DeepResearch \\
\noindent\textbf{Final Vote:} Agent B

\vspace{1em}

\section*{Fine-grained Annotation}
\subsection{Text Spans}

\definecolor{titleblue}{RGB}{0, 0, 200}        
\definecolor{frameblue}{RGB}{0, 0, 200}        
\definecolor{contentbg}{RGB}{230, 230, 255}    

\newcommand{\textspanbox}[3]{%
  \noindent
  \fcolorbox{frameblue}{titleblue}{%
    \begin{minipage}{\dimexpr\linewidth-2\fboxsep-2\fboxrule\relax}
      \vspace{0.4em}
      \textcolor{white}{\textbf{\Large #1}}
      \vspace{0.4em}
    \end{minipage}%
  }%
  
  \vspace{-\fboxrule}
  
  \noindent
  \fcolorbox{frameblue}{contentbg}{%
    \begin{minipage}{\dimexpr\linewidth-2\fboxsep-2\fboxrule\relax}
      \vspace{0.5em}
      
      #2
      
      \vspace{0.5em}
      \textbf{Vote:} #3
      \vspace{0.5em}
    \end{minipage}%
  }%
  
  \vspace{1.5em}  
}

\noindent\textbf{Agent}: Perplexity DeepResearch

\vspace{1em}
\textspanbox{Text Span 1}{%
  \textit{``The inclusion of personal life details in journalist biographies represents a complex intersection of transparency efforts, branding strategies, and ethical considerations in contemporary journalism.''}
}{Upvote}

\textspanbox{Text Span 2}{%
  \textit{``Research indicates that while personal disclosures can enhance audience familiarity with reporters, they yield minimal impact on perceptions of professional credibility and integrity.''}
}{Upvote}

    
    



\subsection{Intermediate Steps}

\noindent\textbf{Agent}: \oursys

\vspace{1em}

\definecolor{stepgray}{RGB}{100, 100, 100}      
\definecolor{stepframe}{RGB}{100, 100, 100}     
\definecolor{stepbg}{RGB}{245, 245, 245}        

\newcommand{\stepbox}[2]{%
  \noindent
  \fcolorbox{stepframe}{stepgray}{%
    \begin{minipage}{\dimexpr\linewidth-2\fboxsep-2\fboxrule\relax}
      \vspace{0.4em}
      \textcolor{white}{\textbf{#1}}
      \vspace{0.4em}
    \end{minipage}%
  }%
  
  \vspace{-\fboxrule}
  
  \noindent
  \fcolorbox{stepframe}{stepbg}{%
    \begin{minipage}{\dimexpr\linewidth-2\fboxsep-2\fboxrule\relax}
      \vspace{0.5em}
      #2
      \vspace{0.5em}
    \end{minipage}%
  }%
  
  \vspace{1.5em}
}

\stepbox{Intermediate Step (Part 1)}{%
  \textbf{Step 1}
  
  \textbf{Thought:}
  
  **Planning a Deep Dive into Journalists' Credibility** Okay, so I'm tackling this interesting question: How do the personal lives of journalists impact how the public perceives their professionalism? I'm already formulating my plan. First, I need to get organized. I'm thinking of structuring this research article in a way that builds a solid argument. My initial approach is to start with a strong **introduction**, setting the stage and clearly defining the scope of my investigation. Then, I'll delve into a thorough **literature review**, laying out the existing studies and relevant theoretical frameworks that inform this complex relationship. Next, since this isn't an original study, it will be a deep dive into an **analysis of existing methodology** and existing research methodologies. I need to focus on what research has already been completed, and what are its shortcomings.
  
  ...
  
  \textbf{ACTION : Plan:}
  
  1. **Introduction:** * Define "journalist biographies" in this context (e.g., "About Me" pages, social media profiles, author bios in articles).
  
  * State the core question: How does the inclusion of personal life details affect public perception of journalist credibility and integrity?
  
  * Briefly outline the importance of the topic (e.g., trust in media, role of journalists in society).
  
  * Provide a roadmap for the article.
}

\stepbox{Intermediate Step (Part 2)}{%
  2. **Theoretical Framework/Literature Review:**
  
  * **Credibility:** Define and discuss established models of source credibility (e.g., expertise, trustworthiness, goodwill).
  
  * **Integrity:** Define journalistic integrity (e.g., objectivity, fairness, accuracy, ethical conduct).
  
  * **Self-Disclosure Theory:** Explore how self-disclosure can impact perceptions and relationship building.
  
  * **Parasocial Interaction Theory:** Discuss how audiences form relationships with media figures and how personal details might foster or hinder this.
  
  * **Impression Management Theory:** How journalists might use personal details to manage their public image.
  
  * Review existing studies (if any) that touch upon personal revelations by professionals (not just journalists) and public perception.
  
  ...
  
  \textbf{Vote:} Upvote
}

\section{Detailed results for Human Annotation}

\begin{figure}[H]
    \centering
    \includegraphics[width=\linewidth]{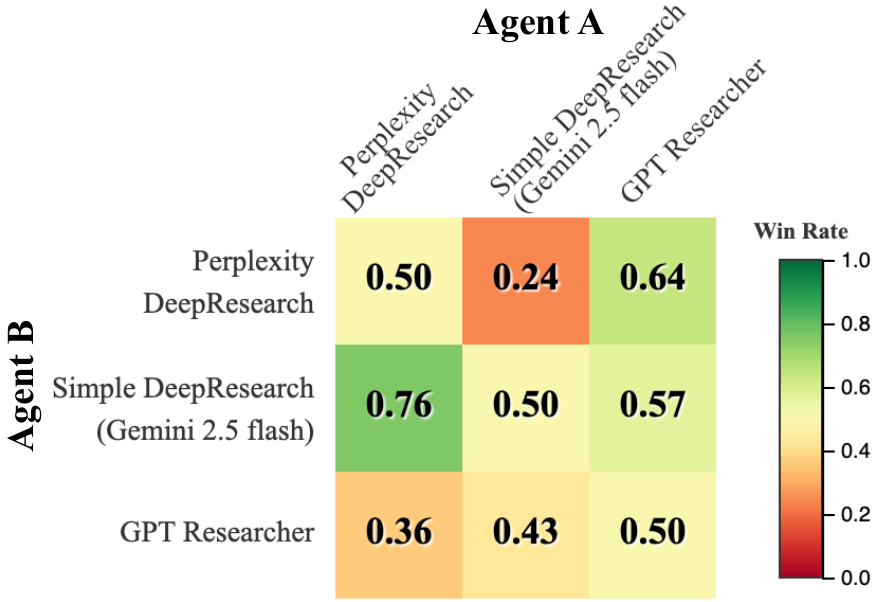}
    \caption{Win Probability of Agent A over Agent B.}
    \label{fig:win_matrix}
\end{figure}

\begin{table}[h]
  \centering
  \caption{
    Upvote/downvote counts for fine-grained annotation. 
    PD: Perplexity DeepResearch, GR: GPT Researcher, SD: \textsc{\oursys} (Gemini 2.5 Flash).
  }
  \label{tab:fine_feedback}
  \small
  \begin{tabular}{lcccc}
    \toprule
    & \multicolumn{2}{c}{\textbf{Intermediate Step}} & \multicolumn{2}{c}{\textbf{Text Span}} \\
    \cmidrule(lr){2-3} \cmidrule(lr){4-5}
    \textbf{Agent} & \textbf{Upvote} & \textbf{Downvote} & \textbf{Upvote} & \textbf{Downvote} \\
    \midrule
    GR & 255 & 33 & 445 & 151 \\
    PR & 161 & 17 & 352 & 36 \\
    SD & 101 & 27 & 271 & 29 \\
    \bottomrule
  \end{tabular}
\end{table}


\clearpage

\begin{table*}[t!]
 \small
 \centering
 \caption{Comparison of \ourplatform with related benchmarks/platforms along key dimensions.} 
\begin{tabular}
{l|c|c|c}
\toprule
\multirow{2}{*}{Benchmark/ Platform}& \multirow{2}{*}{\parbox{2cm}{\centering Real World User Query}} &\multirow{2}{*}{\parbox{2cm}{\centering Long Report}} & \multirow{2}{*}{\parbox{2cm}{\centering Fine-grained Evaluation}}
 \\
\\\midrule
\textbf{\ourplatform} & \cmark&\cmark &   \cmark   \\
\midrule
  Search Arena ~\cite{miroyan2025searcharenaanalyzingsearchaugmented}              & \cmark & \xmark & \xmark \\
  Deep Research Bench ~~\cite{futuresearch2025deepresearchbenchevaluating} & \xmark & \cmark & \xmark \\

Deep Research Gym ~\cite{coelho2025deepresearchgymfreetransparentreproducible} & \xmark & \cmark & \xmark \\
  BrowseComp~\cite{wei2025browsecompsimplechallengingbenchmark}  & \xmark & \xmark & \xmark \\
  GAIA ~ \cite{mialon2023gaiabenchmarkgeneralai} & \xmark & \xmark & \xmark \\
  GPQA ~ \cite{rein2023gpqagraduatelevelgoogleproofqa} & \xmark & \xmark & \xmark \\
  Humanitiy's Last Exam ~\cite{phan2025humanitysexam} & \xmark & \xmark & \xmark \\
\bottomrule 
\end{tabular}
 \label{tab:dataset_comparison}
\end{table*}



\begin{table*}[t!]
  \centering
\caption{Scores of deep research agents on DeepResearchGym using (i) commercial search API and (ii) DeepResearchGym's search API. \textsc{\oursys} (Qwen3-8B) and \textsc{\oursys} (Gemini 2.5 flash) are our baseline agent scaffold integrated with different LLMs. For GPT Researcher with commercial search API, we ran the evaluation using the newly released deep research mode, and other results are extracted from the original paper~\cite{coelho2025deepresearchgymfreetransparentreproducible}.}
\label{tab:benchmarking_results_dsg}
  \renewcommand{\arraystretch}{1.2}
  \resizebox{\textwidth}{!}{\begin{tabular}{l rr rr}
    \toprule
    & \multicolumn{2}{c}{\textbf{Commercial API}} 
    & \multicolumn{2}{c}{\textbf{DRGym API}} \\
    \cmidrule(lr){2-3} \cmidrule(lr){4-5}
    \textbf{System}  
    & Clarity & Insight & Clarity & Insight \\
    \midrule

    Perplexity DeepResearch 
    & 89.5 & 89.3 & -- & -- \\

    gpt4-search-preview         
    & 70.1 & 59.1 & -- & -- \\

    \midrule \midrule
    
    GPT Researcher              
    & \textbf{89.8} & \textbf{90.4} & 83.7 & 78.0 \\

    OpenDeepSearch              
    & 59.2 & 47.0 & 61.5 & 49.5 \\

    HuggingFace-DeepSearch      
    & 57.5 & 48.0 & 58.3 & 52.4 \\

    \midrule

    \textsc{\oursys} (Qwen3-8B)          
    & -- & -- & 75.6 & 78.1 \\

    \textsc{\oursys} (Gemini 2.5 flash)  
    & -- & -- & \textbf{89.4} & \textbf{79.4} \\

    \bottomrule
  \end{tabular}}
\end{table*}

\begin{figure*}[t!]
  \centering
  \includegraphics[width=0.9\textwidth]{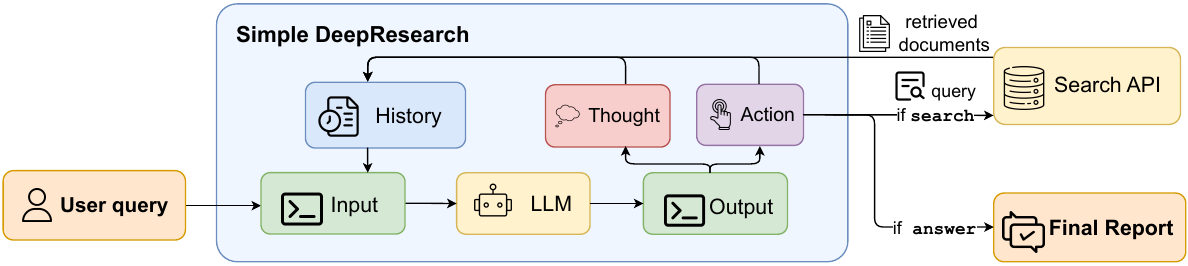}
  \caption{Overview of \textsc{\oursys}, our Baseline agent scaffold.}
  \label{fig:Scaffolding Moc}
\end{figure*}

\end{document}